# Towards a Large Language-Vision Question Answering Model for MSTAR Automatic Target Recognition

David F. Ramirez[12*], Tim L. Overman[2], Kristen Jaskie[12], Marv Kleine[2], Andreas Spanias[1]
[1]SenSIP Center, School of ECEE, Arizona State University, Tempe, Arizona
[2]Prime Solutions Group, Goodyear, Arizona
*dframire@asu.edu

## ABSTRACT

Large language-vision models (LLVM), such as OpenAI's ChatGPT and GPT-4, have gained prominence as powerful tools for analyzing text and imagery. The merging of these data domains represents a significant paradigm shift with far-reaching implications for automatic target recognition (ATR). Recent transformer-based LLVM research has shown substantial improvements for geospatial perception tasks. Our study examines the application of LLVM to remote sensing image captioning and visual question-answering (VQA), with a specific focus on synthetic aperture radar (SAR) imagery. We examine newly published LLVM methods, including CLIP and LLaVA neural network transformer architectures.

We have developed a work-in-progress SAR training and evaluation benchmark derived from the MSTAR Public Dataset. This has been extended to include descriptive text captions and question-answer pairs for VQA tasks. This challenge dataset is designed to push the boundaries of an LLVM in identifying nuanced ATR details in SAR imagery. Utilizing parameter-efficient fine-tuning, we train an LLVM method to identify fine-grained target qualities at 98% accuracy. We detail our data setup and experiments, addressing potential pitfalls that could lead to misleading conclusions.

Accurately identifying and differentiating military vehicle types in SAR data poses a critical challenge, especially under complex environmental conditions. Mastering this target recognition skill may require a human analyst months of training and years of practice. This research represents a unique effort to apply LLVM to SAR applications, advancing machine-assisted remote sensing ATR for military and intelligence contexts.

**Keywords:** synthetic aperture radar, visual question answering, large language vision models, neural networks, machine learning, image captioning, remote sensing, automatic target recognition

## 1. INTRODUCTION

The release of ChatGPT by OpenAI in 2022 opened the world's eyes and marked the beginning of the Artificial Intelligence (AI) era [1]. Other technology companies have developed competing AI products, while academic research and open-source communities have created comparable AI methods. These techniques started as text generation "chatbots" but have progressed into powerful tools for answering questions and performing tasks.

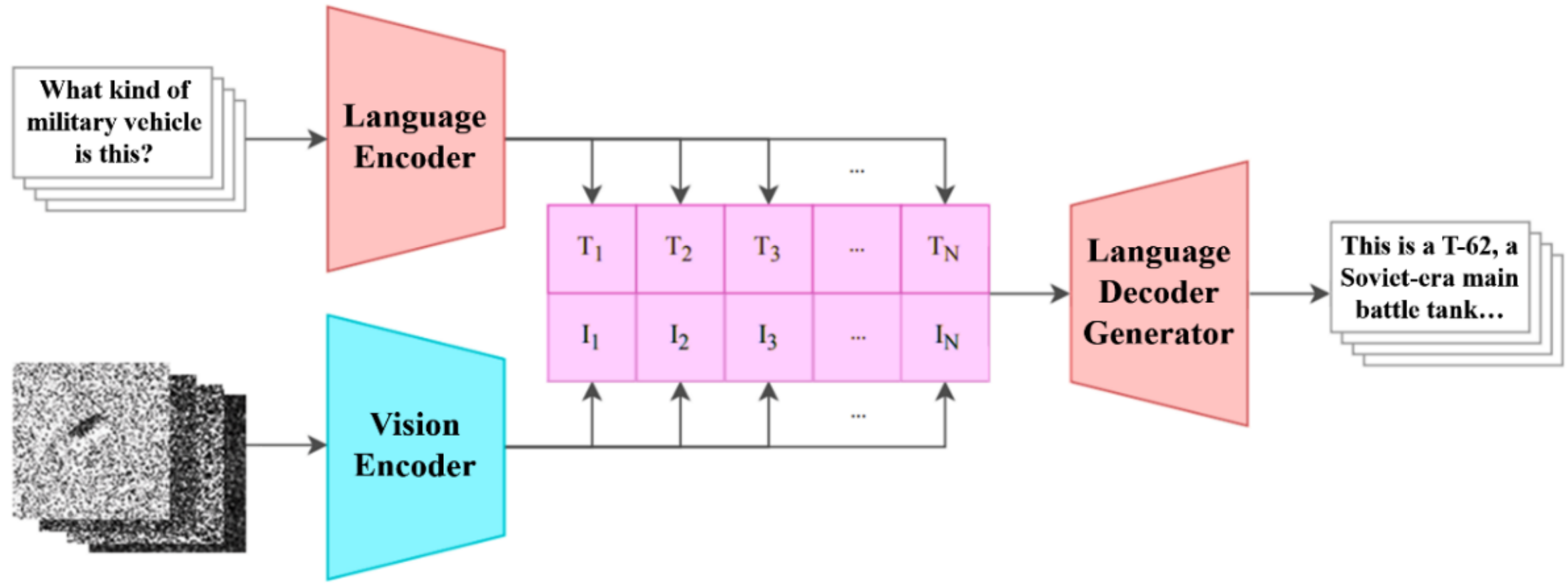

Figure 1. Separately trained language encoder, vision encoder, and language decoder are combined to form a new capability.

Today, language, imagery, and other data types can be provided as input to these methods, but only if they are built and trained in this manner. The expanding capabilities of the so-called large language vision models (LLVM) present an opportunity for intelligence, surveillance, and reconnaissance (ISR) operations and automatic target recognition (ATR) tasks. If these AI methods can be trained for new data domains, such as synthetic aperture radar (SAR), a powerful new capability could be created for traditionally challenging tasks.

We present a feasibility study to fine-tune and evaluate an LLVM for SAR ATR tasks, including target classification, image captioning, and visual question answering (VQA).

### 1.1 Synthetic Aperture Radar

SAR is a highly challenging sensor modality that benefits from ATR algorithms. Unlike typical photography, which uses optics and photosensors, SAR transmits and receives radar from different angles, combining signal processing and geometry to calculate an image [2]. This SAR-formed image has radar-echo energy and timing information encoded into each pixel. The phase timing data can be discarded to keep only a grayscale picture of the radar magnitude. Figures 1 and 2 illustrate the complexity of SAR data, which can be noisy, low resolution, and geometrically distorted. SAR data may be of a large geographic region or centered on a target of interest for ATR. Only a well-trained and experienced eye for detail can differentiate subtle differences between SAR ATR images. A SAR intelligence analyst could be empowered by an AI assistant that executes agentic tasks, like detecting objects, comparing imagery, and automatic label annotation.

### 1.2 MSTAR Dataset

SAR for ATR is heavily influenced by the Moving and Stationary Target Acquisition and Recognition (MSTAR) dataset [3]. This data was produced by the Air Force Research Laboratory, DARPA, and Sandia National Laboratory in 1995 and 1996. For the past three decades, the MSTAR public dataset has served as a key benchmark for academic research [4]. Examples of this dataset are shown in Figure 2. The MSTAR data includes radar collections of various vehicles of interest from multiple locations, along with calculated SAR-formed images and substantial metadata. Many predictive algorithms and machine learning (ML) techniques have been applied to the MSTAR dataset for target detection and recognition [4-8]. The MSTAR dataset serves as a challenging benchmark for ATR algorithms, particularly for the vehicle classification task. To our knowledge, the MSTAR imagery and metadata have not been used for remote sensing image captioning or VQA tasks, which we explore in our research.

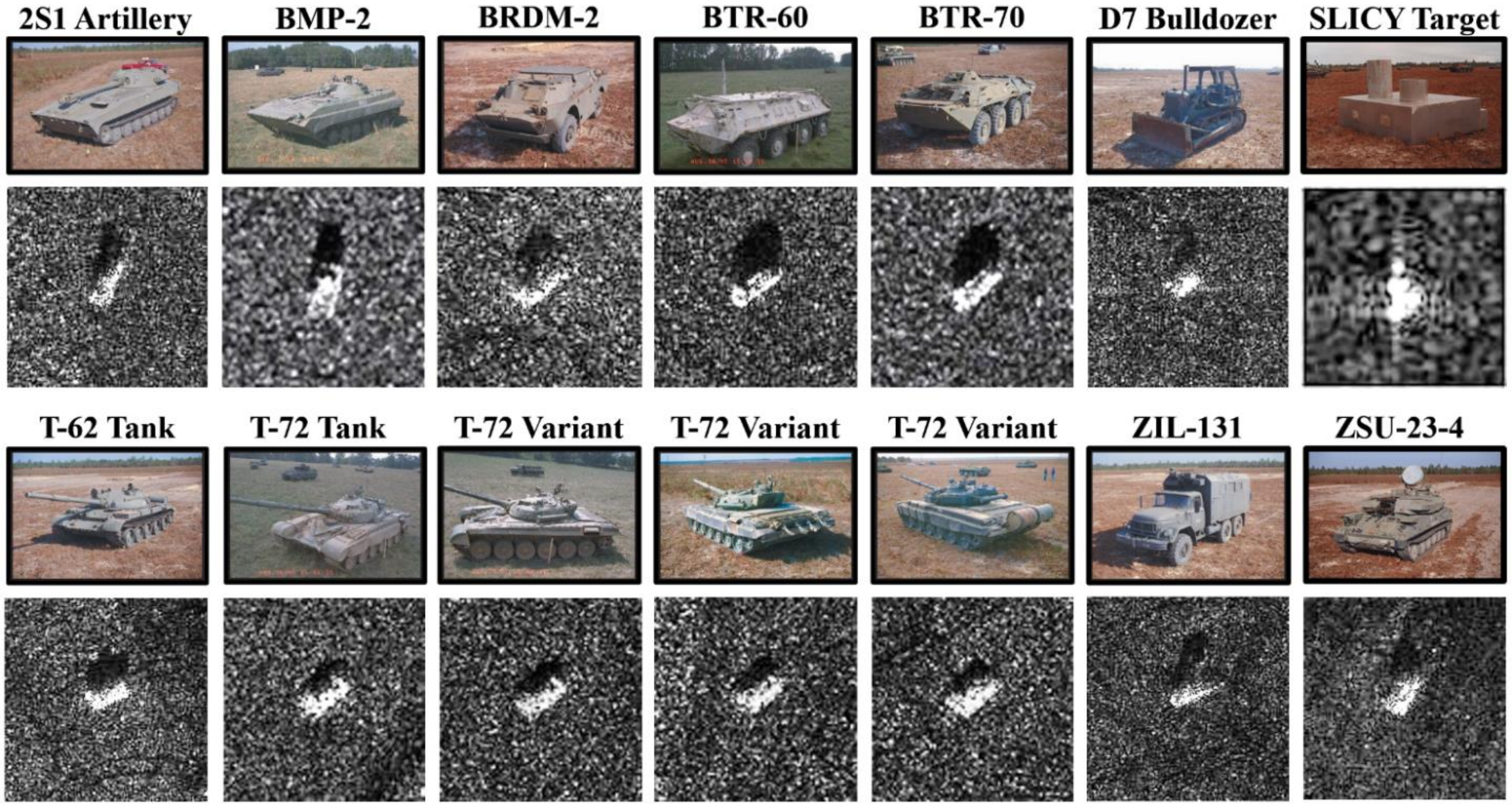

Figure 2. The MSTAR dataset produced by AFRL and DARPA includes military vehicles centered in SAR images [3].

# 2. RELATED WORK

## 2.1 Large Language and Vision Models

An LLVM, like a large language model (LLM), is a neural network based on the transformer architecture [9]. Like most ML algorithms, data is trained into the model to optimize for a task, known as an objective or loss function. Unlike traditional ML and deep learning methods, which are trained on a focused scope of data, an LLM and LLVM are initially trained on a comprehensive set of all available data. A massive quantity of text and imagery is often aggregated by scraping the entire internet [10]. This expansive training data is then trained into an LLM much like an auto-encoder: the model attempts to reconstruct the input. Known as generative pre-training [11], this loosely controlled or semi-supervised training method familiarizes the model with a broad range of subject matters. The generative pre-training process is illustrated in Figure 3. Generatively pre-trained transformers (GPT) are synonymous with LLM and represent state-of-the-art methods for language understanding [10]. A GPT is a causal and recursive system that uses only preceding words or characters, known as tokens, to predict the next single token in an auto-regressive manner. The entire token history, including appended predictions, is then repeatedly ingested until a complete sequence is generated.

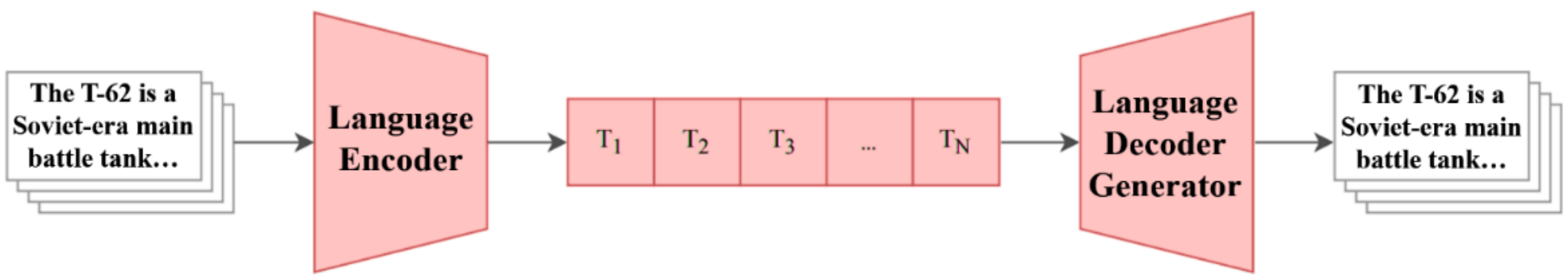

Figure 3. Generative pretraining of a language transformer allows it to learn word associations for follow-on applications.

Over the past few years, commercial and academic researchers have rapidly developed numerous, progressively improved LLMs and expanded their capabilities to include imagery. Large technology companies, such as Meta/Facebook, have generously trained and released powerful GPT LLMs, including the Large Language Model Meta AI (LLaMA) series of models. However, many of these models have licenses with legal use restrictions for commercial and military applications. Other startup companies have released high-performing LLMs without releasing code or reproduction methods, such as Mistral-7B [12]. Researchers have reused, adapted, and extended these models to create entirely open-source LLMs, such as Falcon, Alpaca, and Vicuna. OpenAI's GPT-3 and GPT-4 are examples of closed-source, proprietary LLVMs. Nevertheless, OpenAI has published many technical advancements in academic literature. One principal research advancement is Contrastive Language-Image Pre-training (CLIP) [13], initially published in 2021, a training technique designed to solve large-scale image understanding. Alternative methods, such as ALIGN and BLIP, were also published during this time. These fundamental training techniques produce neural network structures integrated into LLVMs, like the Large Language-and-Vision Assistant (LLaVA) method [14].

## 2.2 Contrastive Language-Image Pre-training

CLIP takes two separate neural network architectures and jointly trains these for unified visual and language understanding [13]. To explain this concept, it can be stated that any input into a neural network is transformed during mathematical processing. An intermediate numeric representation is known as an embedding, and a neural network structured to produce these embeddings is commonly known as an encoder. A GPT typically includes a language encoder and a generative language decoder. A pre-trained language encoder can be combined with a vision encoder from another source, such as an image classifier like a residual neural network (ResNet) or a vision transformer (ViT) [15]. Given two separately trained language and vision encoders, the CLIP method is used to align these to produce similar embedded representations. This CLIP process is illustrated in Figure 4.

CLIP is formulated based on the well-established principles of cosine similarity and cross-entropy. A dataset comprising imagery and text information, such as images accompanied by captions of known objects, is used as input. This data is processed with the encoder models to produce numeric vector embeddings of similar size. Batches of data pairs are learned concurrently utilizing a combined similarity and contrastive loss. For the $\mathrm{I}_i$ and $\mathrm{T}_i$ paired image and text caption embeddings along the matrix diagonal, maximizing the dot product improves this alignment. As in cosine similarity, two vectors are considered similar if their dot product is proportionally large. For elements in the off-diagonal, the dot products $\mathrm{I}_i \cdot \mathrm{T}_j$ are minimized, pointing these toward opposite vector directions. These similarity and contrastive expressions are combined with the standard neural network batch-cross-entropy loss equation for classification. Both the image and text losses are combined to form a symmetric cross-entropy loss over similarity. This is presented in Equation 1, where N represents the total number of paired elements, each of which is dissimilar to the others in the batch. The learned scaling factor $t$ represents the temperature, which modifies the variability of the loss calculation and the magnitude of the gradient. Temperature may also be adjusted at test time to increase the variability of prediction probabilities.

$$L(\mathrm{I},\mathrm{T}) = \left(-\frac{1}{\mathrm{N}}\sum_i \ln\frac{e^{\mathrm{I}_i\cdot\mathrm{T}_i/t}}{\sum_j e^{\mathrm{I}_i\cdot\mathrm{T}_j/t}} - \frac{1}{\mathrm{N}}\sum_j \ln\frac{e^{\mathrm{I}_j\cdot\mathrm{T}_j/t}}{\sum_i e^{\mathrm{I}_i\cdot\mathrm{T}_j/t}}\right)/2 \quad (1)$$

CLIP typically takes substantial computing resources, but the resulting foundation models offer a new predictive capability. For example, the larger of the two proposed CLIP models was trained on 592 GPUs, each with 16GB of Video Random-Access Memory (VRAM), for 18 days [13]. The VRAM total is the greatest limiting factor for training. The resulting unified encoder models are then reused to enable various downstream tasks, such as zero-shot image classification, whereby the model can categorize images into classes it has never explicitly seen before. The substantial learning contained within the language model enables vision understanding even when little image data is available for training. This zero-shot understanding is a powerful new capability that benefits application domains where data is sparse.

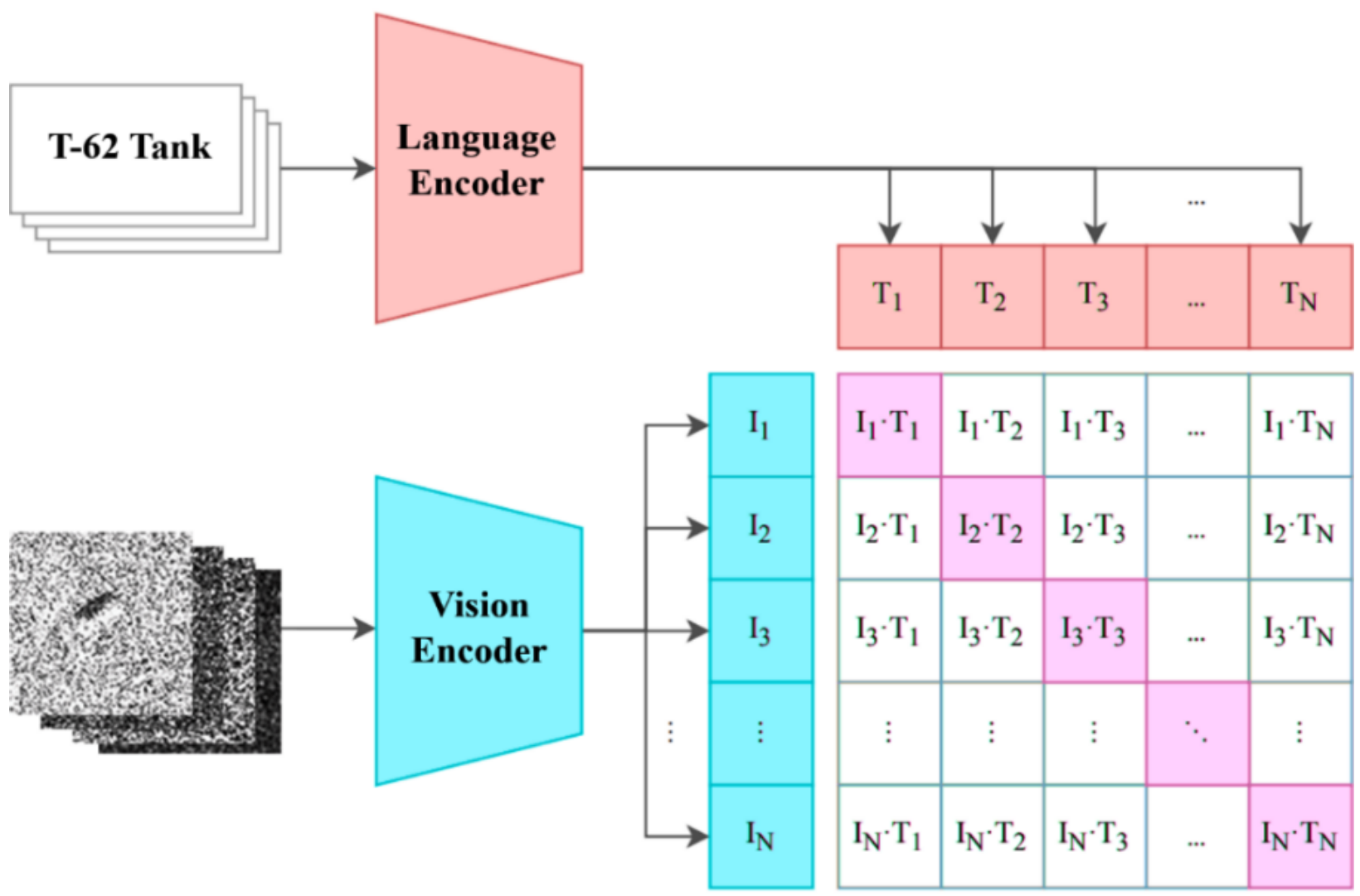

Figure 4. The CLIP training method aligns pre-trained language and vision encoders for joint understanding.

### 2.3 Large Language-and-Vision Assistant

LLVMs combined with modern training techniques enable applied visual understanding with limited computing resources. The LLaVA series of models [14] is a state-of-the-art method first published in 2023, resulting from a collaborative effort between academia and industry. LLaVA is formed from an open-source, state-of-the-art LLM and vision encoder, and numerous possible combinations have been documented. These are then trained for cooperative alignment. A simplified diagram of this setup is shown in Figure 1, and a more detailed diagram is presented in Figure 5. LLaVA adds a linear adapter that projects the vision embedding into the language-embedded space with matching dimensions. With this alignment of the language and vision encoders, each dimension of their embedding vectors captures similar representations. The published LLaVA model was trained on 8 GPUs, each with 80 GB of VRAM, for 18 hours [14]. This LLaVA architecture can be further fine-tuned using Quantized Low-Rank Adaptation (QLoRA) [16] to reduce the memory footprint while training. QLoRA and similar techniques are known as parameter-efficient fine-tuning (PEFT) methods.

### 2.4 Remote Sensing Applications

Applied research has adapted language models for automated remote sensing tasks, including image captioning and VQA. One of the earliest examples of VQA for electro-optical (EO) satellite imagery was the RSVQA method published in 2020 [17]. This early method combined a recurrent neural network (RNN) as a language encoder, the largest ResNet-152 model as the vision encoder, and a dense artificial neural network (ANN) as the language generator. Although this combined model was substantial for its time, with over 60M trained parameters, this does not qualify as an LLVM. Large models are commonly understood as having over a billion trainable neural network parameters.

In 2024, a wave of research was published, including many GPT, CLIP, and LLaVA-based methods for understanding EO satellite imagery. GeoRSCLIP [18] is one of several published methods utilizing CLIP for remote sensing image captioning and VQA tasks. The authors claim to use 1 GPU with 40GB of VRAM for 4 weeks for all training and experiments. RS-LLaVA [19] was published in April 2024 and utilizes the LLaVA architecture for automated remote sensing. Training required two GPUs, each with 48 GB of VRAM, to be run for up to 19.4 hours. Similar EO satellite imagery, image captioning, and question-answering methods are advancing rapidly. New benchmark datasets are being rapidly published to support this research. Recent transformer-based LLVM research has shown substantial improvements for geospatial perception tasks. These methods can be repurposed for SAR imagery, perhaps enabling an uncanny familiarity with data it has never seen before.

### 2.5 SAR Applications

Open SAR research has recently begun to explore language models for ATR. During the recent 2024 Computer Vision and Pattern Recognition (CVPR) Workshops, the winners [20] of the Perception Beyond the Visible Spectrum (PBVS) 2024 Multi-Modal Aerial View Image Challenge for SAR Classification presented the first-place solution. Guo et al. [20] fine-tuned a general-purpose CLIP-aligned ViT-Base with 86M parameters to perform vehicle classification, leveraging novel language information. The radar scattering characteristics obtained from SAR images were converted into textual descriptions, serving as language context paired with the image as input. This winning solution was trained using PEFT on a single GPU with 24GB of VRAM.

Recently, researchers have created SAR-specific foundation models for image embedding encoders. In January 2025, the SARATR-X [21] foundation models were released for MSTAR SAR ATR, utilizing pretraining, target classification, and object detection target labels, but without image captioning or VQA-associated language data. This family of vision embedding models has up to 89M neural network parameters and was trained on 8 GPUs with 24GB of VRAM each.

Another substantial achievement came from Wang et al. [22] in August 2024. These researchers explored SAR ATR utilizing a language model and other advanced generative SAR data methods. MSTAR, SAMPLE, civilian vehicles, generative diffusion, 3D models, and small language models were used for zero-shot classification. Utilizing a downsized 110M-parameter BERT language model and the 86M-parameter ViT-Base vision encoder, their language-vision model was formed and fine-tuned on a single GPU.

LLVM techniques were once inaccessible for most researchers due to their computational demands. However, we now demonstrate that recent advances can make LLVM available for new applications, such as SAR ATR. Foundation models can be leveraged as a powerful tool for remote sensing and SAR, as demonstrated by the winning solution from the referenced CVPR workshop. The substantial pretraining of an LLM may possess deep, latent knowledge that is beneficial to the ATR domain. Even before application-specific training, the LLVM may already understand language and vision tasks, such as analysis, classification, and object detection. Research shows that predictive accuracy improves with the size of a foundation model [10], as larger models can learn, retain, recall, and re-train with greater performance. This research effort presents a first step towards unlocking the power of LLVMs for SAR.

## 3. METHODS

The current state-of-the-art GPT, LLM, and LLVM methods can be adapted to comprehend complex remote sensing tasks. We develop an LLVM for SAR ATR tasks, including classification, image captioning, and VQA, on the MSTAR dataset.

### 3.1 Experiment 1: ATR Classification

We first tested the LLVM method on the traditional MSTAR SAR ATR task of vehicle identification [3], allowing for unrestricted text generation. We chose to increase the difficulty of this evaluation by increasing the number of potential vehicle target types while limiting the amount of SAR image data for training.

We form semi-balanced training and test datasets for our experiment by combining several MSTAR-published datasets. This dataset expands the standard MSTAR Mixed Targets dataset to include as many vehicle categories as possible. The data for Experiment 1 is shown in Table 1. We follow the predefined training/test split convention regarding the 17/15-degree radar collection angle separation. For the 11 vehicle classification categories, each type has between 233-299 training and 194-274 evaluation examples, creating a relatively class-balanced experiment. We chose not to use all the available MSTAR examples for this experiment to maintain a balanced distribution of target class categories through an undersampling strategy. This combined dataset comprises 3,112 training images and 2,778 validation images, each labeled with a vehicle type. Table 1 lists the serial number for reproducibility, but it is not used as a label or feature in the experiment. We utilize SAR images from the 1995 Collect 1-1 at Redstone Arsenal and the 1996 Collect 2-1 at Eglin Air Force Base, as noted in the MSTAR header metadata. The geographically distinct and separately gathered data may skew the results, which will be discussed in the Results section.

We chose to deviate from the standard class-label structure for military vehicle types. To maximize the pre-trained LLVM's latent potential, we converted the vehicle target labels to follow Wikipedia's military vehicle naming conventions, as Wikipedia is a common data source for generative pre-training. For example, we use the target label "BMP-2" instead of "BMP2" or "BMP_2." This language tokenization consideration is discussed in the Machine Learning Training section.

When available, we utilized the JPEG image samples from the MSTAR/IU Mixed Targets and T-72 Variants public datasets [3]. Some Collection 1 MSTAR targets do not include JPEG sample images but only provide raw binary magnitude and phase results. For these BMP-2 and BTR-70 vehicle targets, we generated these images using the "mstar2jpeg" executable provided by the AFRL Sensor Data Management System [3]. In future experiments, we will use these helper functions to create lossless TIFF images for all SAR examples.

Table 1. The MSTAR datasets are combined to capture a challenging quantity of vehicle target types for ATR.

**Experiment 1: ATR Classification**

| **Vehicle** | | **Image Counts** | | **Collection & Scene** |
|---|---|---|---|---|
| **Target Type** | **Serial Number** | **Training (17°)** | **Test (15°)** | |
| 2S1 | B01 | 299 | 274 | 2-1 |
| BMP-2 | C21 | 233 | 196 | 1-1 |
| BRDM-2 | E-71 | 298 | 274 | 2-1 |
| BTR-60 | K10YT7532 | 256 | 195 | 1-1 |
| BTR-70 | C71 | 233 | 196 | 1-1 |
| D7 | 92V13015 | 299 | 274 | 2-1 |
| SLICY | SN-1 | 298 | 274 | 2-1 |
| T-62 | A51 | 299 | 273 | 2-1 |
| T-72 | A64 | 299 | 274 | 2-1 |
| ZIL-131 | E12 | 299 | 274 | 2-1 |
| ZSU-23-4 | D08 | 299 | 274 | 2-1 |
| | | 3112 Targets | 2778 Targets | |

Table 2. These simple text captions describe qualitative details within the MSTAR imagery.

**Experiment 2: ATR Image Captioning**

| **Vehicle** | **Text Caption** |
|---|---|
| 2S1 | "2S1, a tracked amphibious self-propelled howitzer artillery" |
| BMP-2 | "BMP-2, a tracked amphibious infantry fighting vehicle" |
| BRDM-2 | "BRDM-2, a wheeled amphibious armored scout car" |
| BTR-60 | "BTR-60, a wheeled amphibious armored personnel carrier" |
| BTR-70 | "BTR-70, a wheeled amphibious armored personnel carrier" |
| D7 | "D7, a tracked medium tractor bulldozer" |
| SLICY | "SLICY, a simple geometric-shaped static target" |
| T-62 | "T-62, a tracked medium main battle tank" |
| T-72 | "T-72, a tracked medium main battle tank" |
| ZIL-131 | "ZIL-131, a wheeled general-purpose truck" |
| ZSU-23-4 | "ZSU-23-4, a tracked self-propelled anti-aircraft weapon" |

### 3.2 Experiment 2: ATR Image Captioning and VQA

Image captioning is a new domain of research for the MSTAR dataset. Instead of inferring only an ATR vehicle type, the predictive method must generate an extended text description of the image. This longer text generation is typically trained and evaluated against several possible captions. In this feasibility study, we created only a simple text caption of the military vehicle's type and qualities. Table 2 presents the captions assigned to each vehicle target category and the SLICY simulated target. We apply these image captions to the datasets formed in Experiment 1.

Image captioning can be extended into VQA if more details are known or by asking specific questions. After training the model for ATR vehicle classification and image captioning, we probe the model to answer questions that extend beyond its original training. This begins to test the model's ability to answer questions outside the scope of how it was trained, known as zero-shot prediction. We explore zero-shot prediction and VQA for SAR ATR in our future work.

### 3.3 Machine Learning Training

As discussed in the Related Work section, we utilize pre-trained foundation models refined through CLIP alignment, adapters, instruction tuning, and PEFT.

We explored several LLVM types during our research, but this section focuses on the LLaVA-NeXT model [23]. Built to improve upon the LLaVA methods published in 2023 [14] and 2024 [24], LLaVA-NeXT utilizes state-of-the-art language and vision foundation models optimized for image captioning and VQA. We train the LLaVA-NeXT v1.6 Mistral-7B LLVM for SAR ATR. This neural network combines the Mistral 7.3B-parameter language model (version 0.2), the 305.5M-parameter ViT-Large Patch-14 336-pixel vision encoder, and a 21M-parameter projector extension for alignment. Each of these elements was trained progressively by several different researchers. We discuss the preceding foundation-model ML training computational requirements in the Conclusions section.

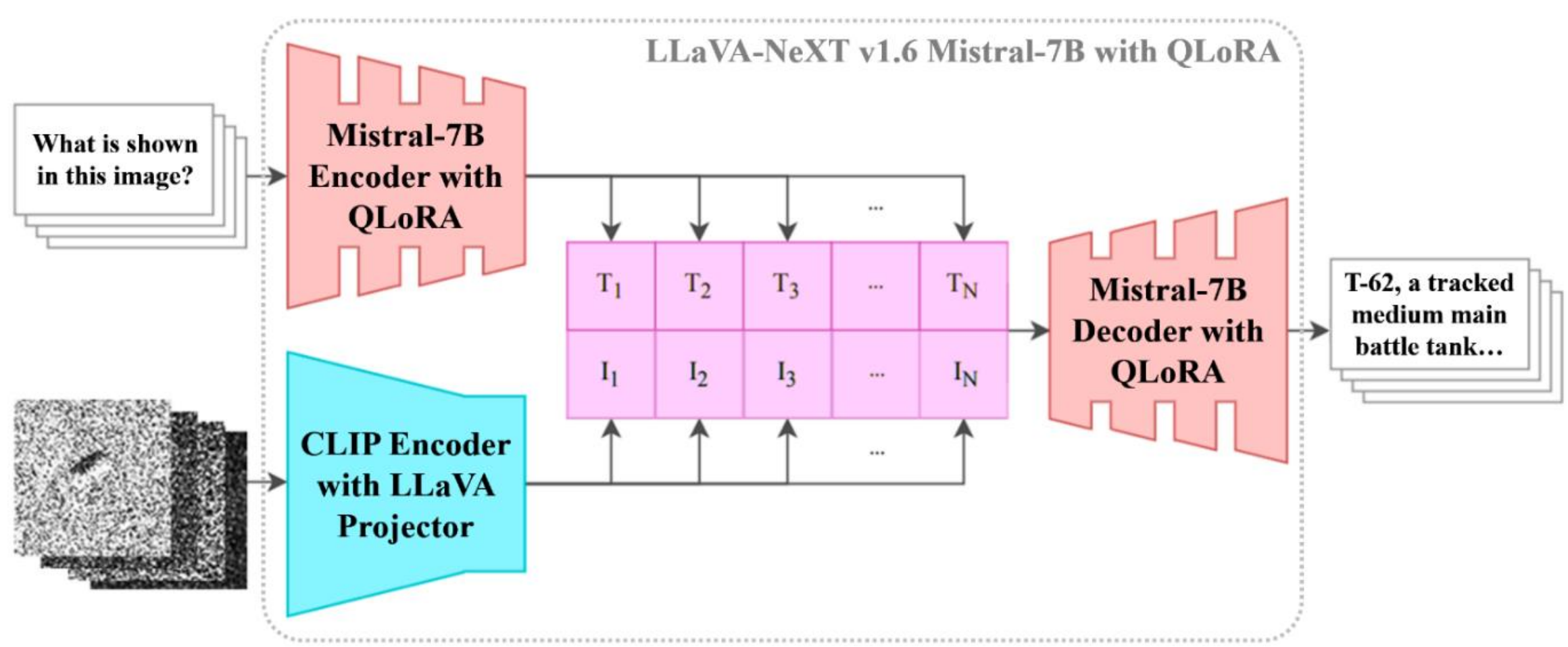

Figure 5. LLaVA combines a fine-tuned LLM encoder and decoder pair with a CLIP-aligned vision encoder and connector.

During training, VRAM must be allocated for all the neural network parameters, as well as their calculated gradients, backpropagation math, and the gradient descent optimizer state. This equals approximately 8 times the original model size in memory [25]. Typically, the LLaVA-NeXT Mistral-7B LLVM requires approximately 30GB of memory to load the model, and a minimum of 240GB of VRAM for training with $batch\ size = 1$. The developers of LLaVA-NeXT required 640GB of GPU VRAM [23] and 20 hours of training. Renting the minimum viable machine from a cloud computing service might cost $41 per hour at the time of writing, totaling over $800 to train a single model. This training requirement is unfeasible for most research activities. Recent advances in PEFT have made retraining much more accessible.

This memory footprint can be reduced by freezing parameters to be non-trainable, adding a limited number of new trainable parameters, or by adjusting the numeric precision of calculations and numbers in memory. We utilize 4-bit Normalized Floating-Point (NF4) quantized numeric representation for the model's parameters and 16-bit Brain Floating-Point (BF16) precision for calculations. These techniques reduce the model footprint in memory from approximately 30GB to 3.8GB. We inject QLoRA [16] trainable neural network weights into the language model encoder and decoder, adding only 4.2M trainable parameters. LoRA parameters were added to every Q and V projection transformer layer in the language encoder and decoder. We utilize the community-accepted starting LoRA values of $alpha = 16$, $dropout = 5\%$, and $rank = 8$. These values can be optimized through hyperparameter tuning, which we leave for future work. These LoRA hyperparameters result in a customized model with a total of 0.0554% trainable parameters. Utilizing the gradient checkpointing method for improved memory efficiency, we determine that the minimum VRAM required to train this adapted model is 5.68GB. PEFT was conducted on a single GPU with 24GB of VRAM. A maximum of $batch\ size = 8$ parallel training examples is possible, resulting in approximately 21GB of VRAM use.

The ViT-Large Patch-14 image-transformer model sub-samples an image into many smaller embeddings [15]. The Patch-14 model-variety samples images through windowing, creating 14x14 pixel patches before processing. The encoder will then produce an embedding for each of these sub-image patches, resulting in many vision embeddings for each image. For the standard 128x128 pixel MSTAR image, the ViT-L Patch-14 encoder will produce 100 total patch embeddings per image. During ML training, an LLVM and its generator will learn relationships that link image embeddings, such as the associations between different words and sentences in proximity.

We train the LLaVA-NeXT Mistral-7B LLVM through QLoRA for only two epochs of the training dataset. With $batch\ size = 8$ this is a total of 778 training steps. We use the PyTorch implementation of a fused AdamW gradient descent optimizer with constant $learning\ rate = 0.0002$ and the default values of $\beta_1 = 0.9$, $\beta_2 = 0.999$, and $weight\ decay = 0.01$. This was initialized with $warmup\ steps = 50$ to stabilize momentum and adaptive learning rates. We periodically evaluate the model's performance against the Test dataset, although this is computationally expensive, slowing the training process. During Experiment 1, we evaluated and saved a model checkpoint every 100 training steps. We conducted only a single training run for this experiment. To track performance consistency, we score every checkpointed model saved after the first epoch.

The Mistral LLM requires the use of a specialized tokenizer to convert words to prediction categories. The Mistral tokenizer is based on similar SentencePiece Byte-Pair Encoding methods. As with all machine learning methods, a predictive model generates numbers; it does not directly generate text or other data types. For most LLMs, full words are linked to numeric token identifiers (IDs) in the language model's dictionary. When a word is not in the vocabulary of the language model and matching tokenizer, it must be split into sub-words or characters. In Experiments 1 and 2, the MSTAR vehicle type labels must be divided into sub-words and characters before tokenization, which is suboptimal. Table 3 shows examples of how a string of words is separated and converted into token IDs. When an LLM generates text, it performs token prediction from among all the token types in its vocabulary. Tokenizers with larger vocabulary sizes may need to split words less often; however, including more token IDs to predict against will make the task more challenging. Considering two tokenizers, such as Mistral and LLaMA-1, even when their vocabulary sizes are the same, the token IDs may not match. Similarly, even when most of the words and tokens match, some tokens may not. The tokenizers for GPT-3.5, GPT-4, and LLaMA-3 appear to be highly compatible, although LLaMA-3 expands the vocabulary size to introduce new agentic capabilities. This type of extended vocabulary can be adapted to represent new concepts and actions. In our future work, we intend to represent the MSTAR target vehicle types using single special tokens.

Table 3. Tokenization methods and examples show how words are split into numeric identifiers for token classification [26].

| **Tokenizer** | **Vocab. Size** | **Token Splitting Example [Token IDs]** |
|---|---|---|
| GPT-2, GPT-3 r50k-base | 50,257 | This is a simple geometric-shaped static target, a SLICY.<br>[1212, 318, 257, 2829, 38445, 12, 16760, 9037, 2496, 11, 257, 12419, 2149, 56, 13] |
| GPT-3.5, GPT-4 cl100k-base | 100,256 | This is a simple geometric-shaped static target, a SLICY.<br>[2028, 374, 264, 4382, 53584, 35831, 1118, 2218, 11, 264, 17216, 39364, 13] |
| GPT-4o o200k-base | 199,997 | This is a simple geometric-shaped static target, a SLICY.<br>[2500, 382, 261, 4705, 82570, 53876, 1633, 3783, 11, 261, 336, 11720, 56, 13] |
| LLaMA-1, LLaMA-2 | 32,000 | This is a simple geometric-shaped static target, a SLICY.<br>[1, 910, 338, 263, 2560, 26224, 29899, 845, 10501, 2294, 3646, 29892, 263, 27146, 2965, 29979, 29889] |
| LLaMA-3 | 128,256 | This is a simple geometric-shaped static target, a SLICY.<br>[2028, 374, 264, 4382, 53584, 35831, 1118, 2218, 11, 264, 17216, 39364, 13] |
| Mistral | 32,000 | This is a simple geometric-shaped static target, a SLICY.<br>[3260, 349, 264, 3588, 28094, 28733, 21501, 1062, 2718, 28725, 264, 20375, 1604, 28802, 28723] |

# 4. RESULTS

## 4.1 Experiment 1: ATR Classification

Our QLoRA-trained LLaVA-NeXT Mistral-7B LLVM was able to accurately predict most vehicle categories with only limited training time and without hyperparameter tuning. During the 778 training steps, we took model snapshots every 100 training steps. The results are shown in Table 4. During the initial few hundred steps, we found the model was unable to generate all the vehicle categories. Not until the checkpoint at 300 steps was the model able to generate all 11 vehicle class labels. Although this was not the strongest-performing model, it nonetheless achieved 91.6% accuracy and a class-balanced accuracy of 91.0%. The training at 300 steps is still before the first epoch, with not all training examples having been seen yet by the model. After the first epoch, we aggregate the scores of all saved models to measure consistency.

Table 4. A single training run was monitored, with each model checkpoint evaluated and scored towards the aggregate.

**Experiment 1 Test Results**

| Model Checkpoint | Accuracy | Class-Balanced Accuracy | Class-Balanced Precision | F1-Score |
|---|---|---|---|---|
| Step 400 | 96.25% | 95.37% | 96.56% | 95.23% |
| Step 500 | 92.62% | 92.48% | 93.51% | 91.94% |
| **Step 600** | **98.16%** | **97.70%** | **97.95%** | **97.71%** |
| Step 700 | 94.99% | 93.90% | 95.68% | 93.69% |
| Step 778 | 96.76% | 96.22% | 97.22% | 96.41% |
| Mean ± σ | 95.76% ± 2.09% | 95.13% ± 2.02% | 95.58% ± 1.71% | 95.17% ± 2.26% |

After 400 training steps and completing the first epoch, the model produces high-quality ATR predictions. During the training run, checkpoint 600 was the highest-performing model, scoring the best on the Test set. The confusion matrix of this model is shown in Figure 6. It should be noted that the results on the Test dataset may not generalize to a third, blind dataset. Cherry-picking the best-performing model often does not translate into performance when tested on new data, even when the data is drawn from the original distribution. In Experiment 2 and future work, we conduct several training runs in a Monte Carlo simulation to measure variability.

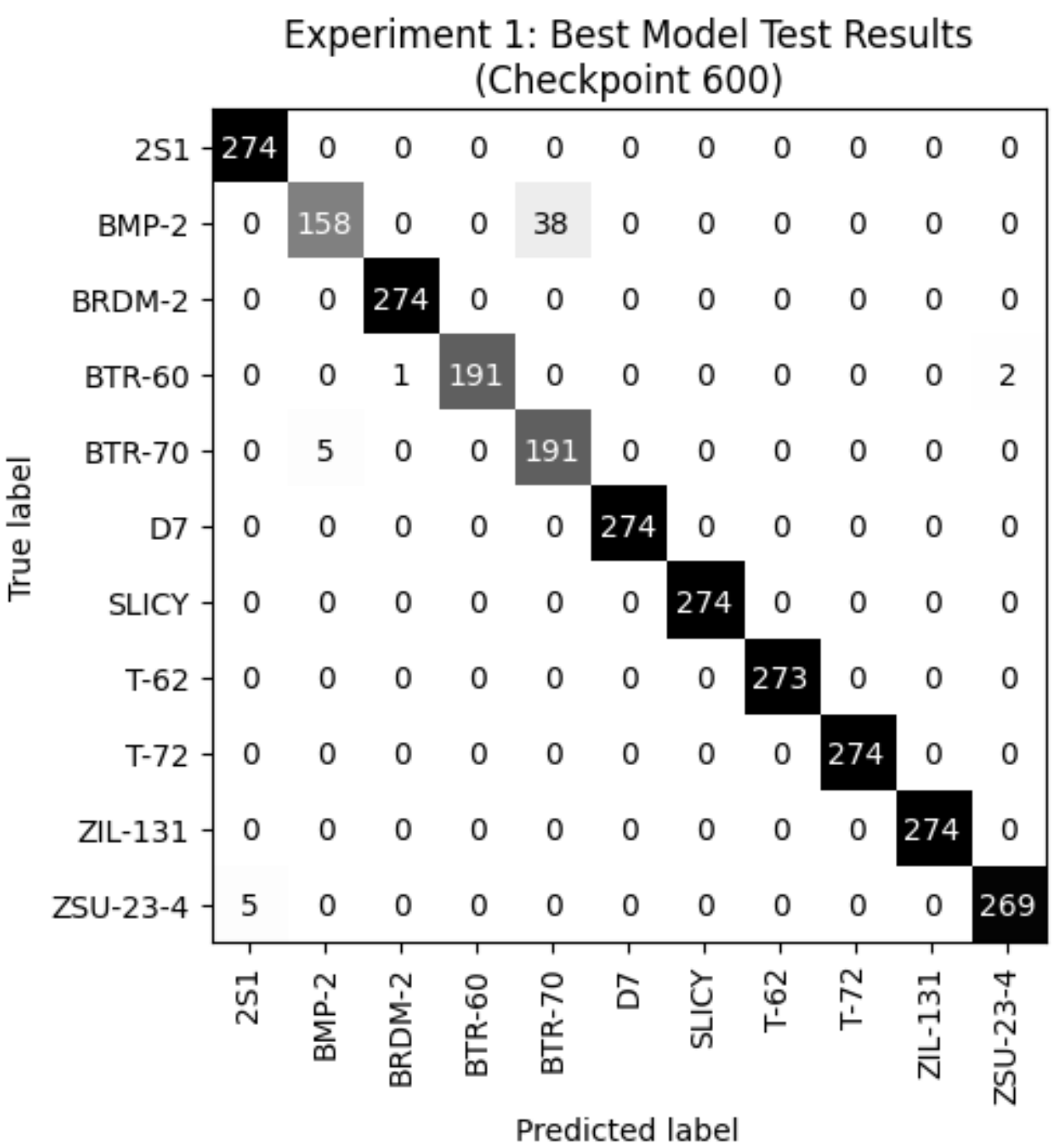

Figure 6. The best-performing model in Experiment 1 exhibits confusion between the BMP-2 and BTR-70 vehicles.

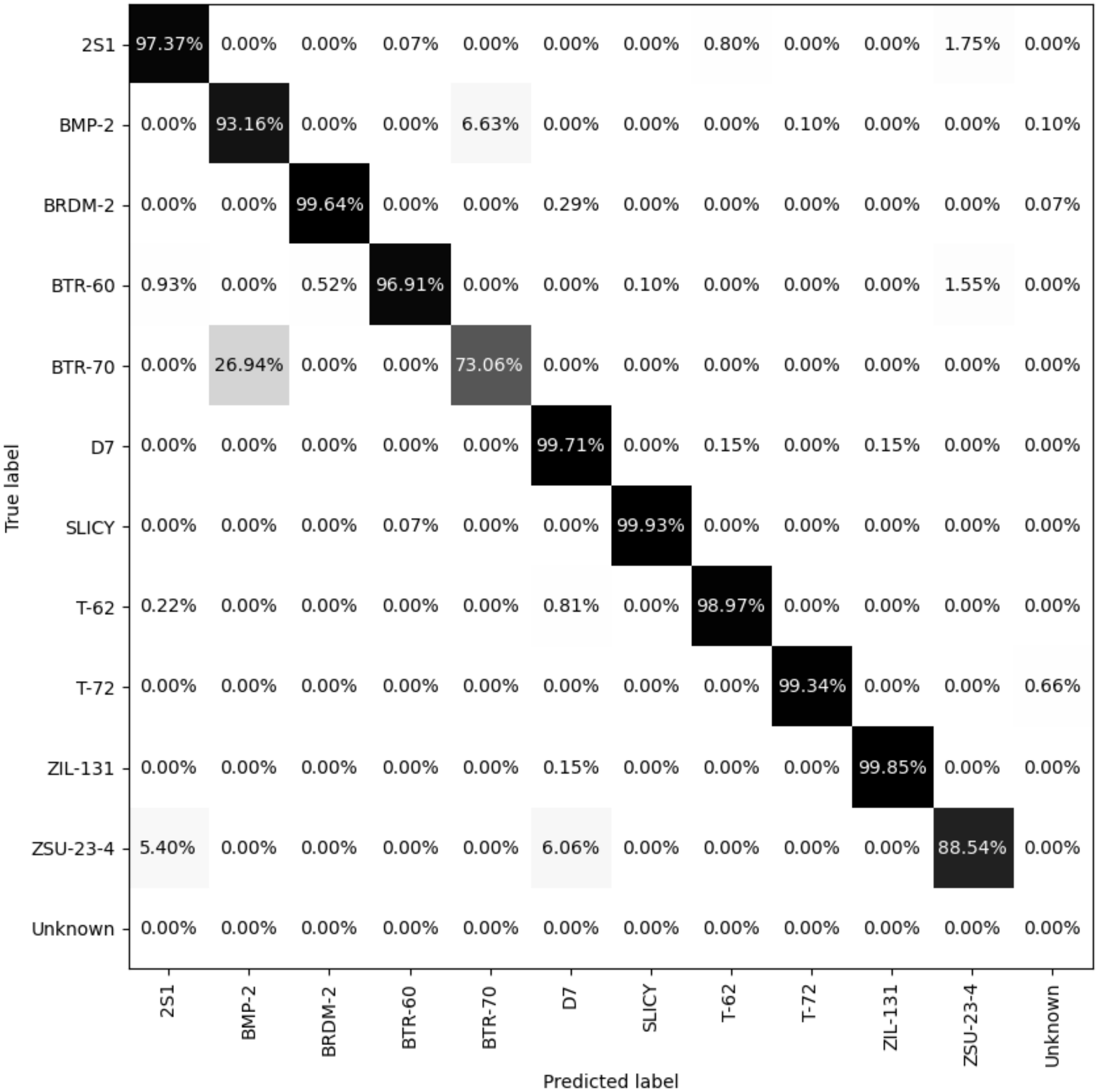

Figure 7. All the model checkpoints saved during Experiment 1 are aggregated and normalized by the true vehicle class.

Upon completing training at step 778, we observed a hallucination phenomenon in the evaluation results. The LLVM-generated predictions included erroneous target types, such as 'D-214', 'D-72', 'D-73', 'DMP-2[/INST]', and 'DMP-28'. The model seems to be mixing the letter tokens of the D7, T-72, and BMP-2 target types. This may be a sign of overfitting or unstable training. Even with these errors, this checkpoint 778 produced an overall accuracy of 96.76%. In Figure 7, these hallucinated vehicle categories are combined into the "Unknown" class.

In Figure 7, the BMP-2 and BTR-70 vehicle types are the most significant sources of errors. We speculate this is due to data bias in the image formation process. These vehicle types are two of the three vehicles SAR imaged in Collect 1, as noted in Table 1. Additionally, these two vehicle types were the only images generated using the "mstar2jpeg" executable provided by the AFRL Sensor Data Management System. It is possible that the trained LLVM learned the JPEG compression scheme associated with this method, or there is some other source of data bias related to the Redstone Arsenal Collect 1 compared to the different captures. We propose solutions to this issue in the Future Work section.

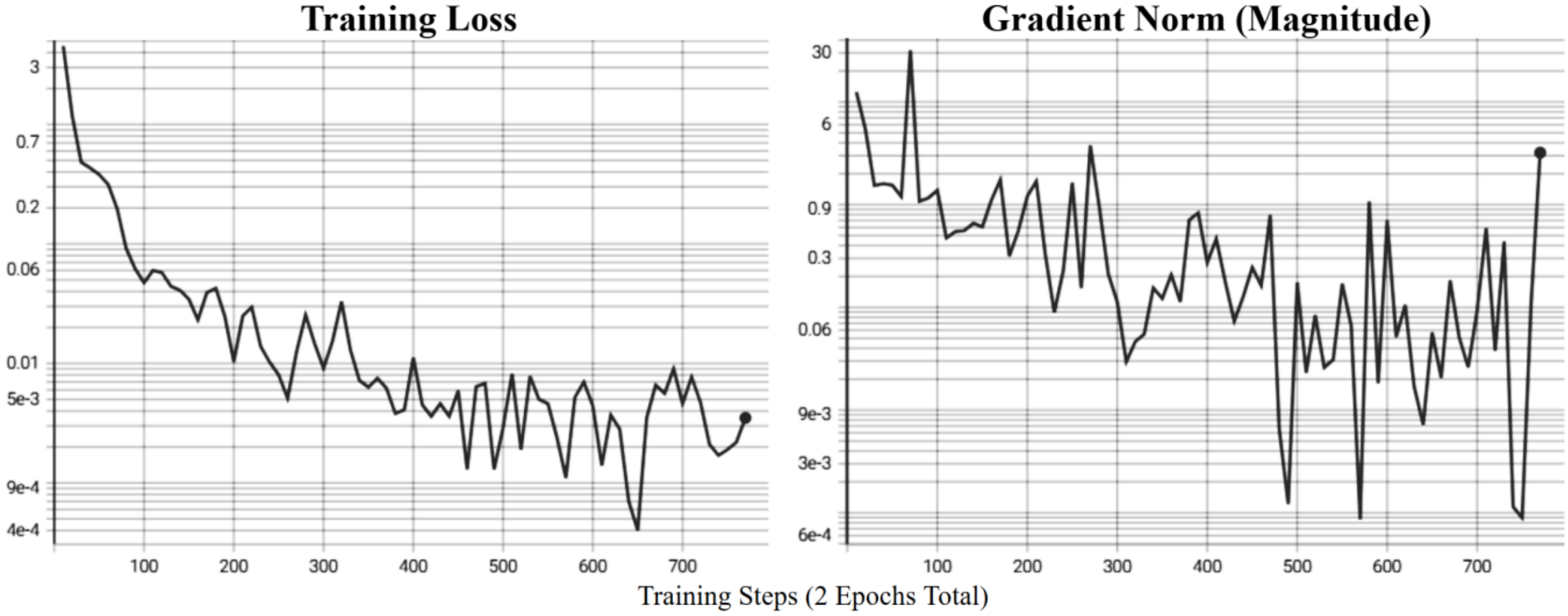

Figure 8. The training loss and gradient magnitude of Experiment 1 are recorded at 10-step intervals on a logarithmic scale.

In Table 4, we observe a substantial change in the evaluated metrics between checkpoints. As noted by the inconsistent training loss and gradient norm magnitude plots shown in Figure 8, the machine learning optimization process seems to be overshooting during training. This may be due to the small quantity of trainable parameters or a learning rate that is too large. This process may also benefit from a learning rate scheduler or a larger batch size to stabilize these fluctuations. The gradient manifold may take the form of a saddle point, with sharp edges in some directions but a slowly descending ideal path that might be followed.

Figure 8 illustrates the variability of the learning process, which can be described using a metaphor. If we think of machine learning optimization as a physical landscape with mountains and valleys, then we want to step downwards towards a higher-performing model. However, reaching the bottom would represent an overfit machine learning model, which only understands the training data. We must use our best judgment to find a comfortably low campsite that avoids hazards. Continuing the metaphor, a maximally low point would be a flooding hazard. The learning rate and optimizer determine our walking pace, serving as a compass to guide us on our journey, showing us where we have been and where we should travel next. The loss metric and plot in Figure 8 represent our elevation on the mountainous landscape. Each calculated loss gradient on a batch of images represents a new shifting landscape. The gradient norm is the slope of the current shifting landscape, indicating whether it is steep or flat. In Figure 8, we observe an increasingly steep footing as the optimizer races us towards the bottom. An adaptive learning rate scheduler could adjust the walking pace to slow down or accelerate. The Adam and AdamW optimizers have adaptive learning rates, which significantly speed up progress down the mountain, but may dangerously stumble into a deep crevasse. Perhaps the slower and steadier approach of stochastic gradient descent (SGD) would lead to a safer stopping point. Each batch of data indeed reveals a new gradient and a shifted landscape that must be navigated. For these graphs, each collection period on these lines represents 10 optimization steps, with batches of 8 examples each, totaling 80 examples per point on the lines. A larger batch size would reveal a more comprehensive map of the environment before we take each step. This creative language was not generated using LLMs. Regardless, an analogy helps to describe these concepts.

### 4.2 Experiment 2: Image Captioning and VQA

We trained an LLVM identical to Experiment 1, but for image captioning labels using the descriptive captions shown in Table 2. This method was trained during three different iterations from the same foundation model starting point, but with randomized training data shuffling. These results are shown in Table 5. When evaluated against the Test dataset, the predictions precisely match the target label phrases word for word. Although the generated image captions are not always accurate, they almost always match the incorrect captions exactly. This is shown in Figure 9. We speculate that the simplicity of this caption template strategy may be too predictable, resulting in a language decoder that cannot produce more diverse outputs or answer other questions. This has also been briefly tested and shown to be the case. Note that we did not utilize probabilistic sampling to generate test outcomes; however, this approach would yield more diverse outcomes. These topics, including overfitting, data variance, and randomization, are explored in the Future Work section.

Table 5. Models trained for exactly two epochs without hyperparameter tuning, stepwise evaluation, or checkpointing.

**Experiment 2 Test Results**

| Final Model | Accuracy | Class-Balanced Accuracy | Class-Balanced Precision | F1-Score |
|---|---|---|---|---|
| Run 1 | 97.95% | 97.63% | 97.88% | 97.70% |
| **Run 2** | **98.06%** | **97.91%** | **97.89%** | **97.87%** |
| Run 3 | 96.98% | 96.91% | 96.90% | 96.71% |
| Mean ± σ | 97.66% ± 0.59% | 97.48% ± 0.52% | 97.43% ± 0.57% | 97.43% ± 0.63% |

We attempted to use these trained models to answer several simple zero-shot VQA queries. For example, we prompted the models with “Is this vehicle wheeled or tracked?” and “Is this vehicle amphibious?” which are both target qualities trained during Experiment 2, as shown in Table 2. Very rarely did the generated responses vary from the trained captions, whether evaluated against the Training or Test sets. Most certainly, the model has not forgotten how to answer such questions, since these parameters are frozen. We speculate that the added QLoRA parameters have shifted the responses towards a new learned behavior. Further research is needed to fully leverage the capabilities of the pre-trained LLM in conjunction with the newly trained SAR ATR domain. In a more complex setup, a “mixture of experts” technique can be employed to combine separately trained LLMs to extend their capabilities.

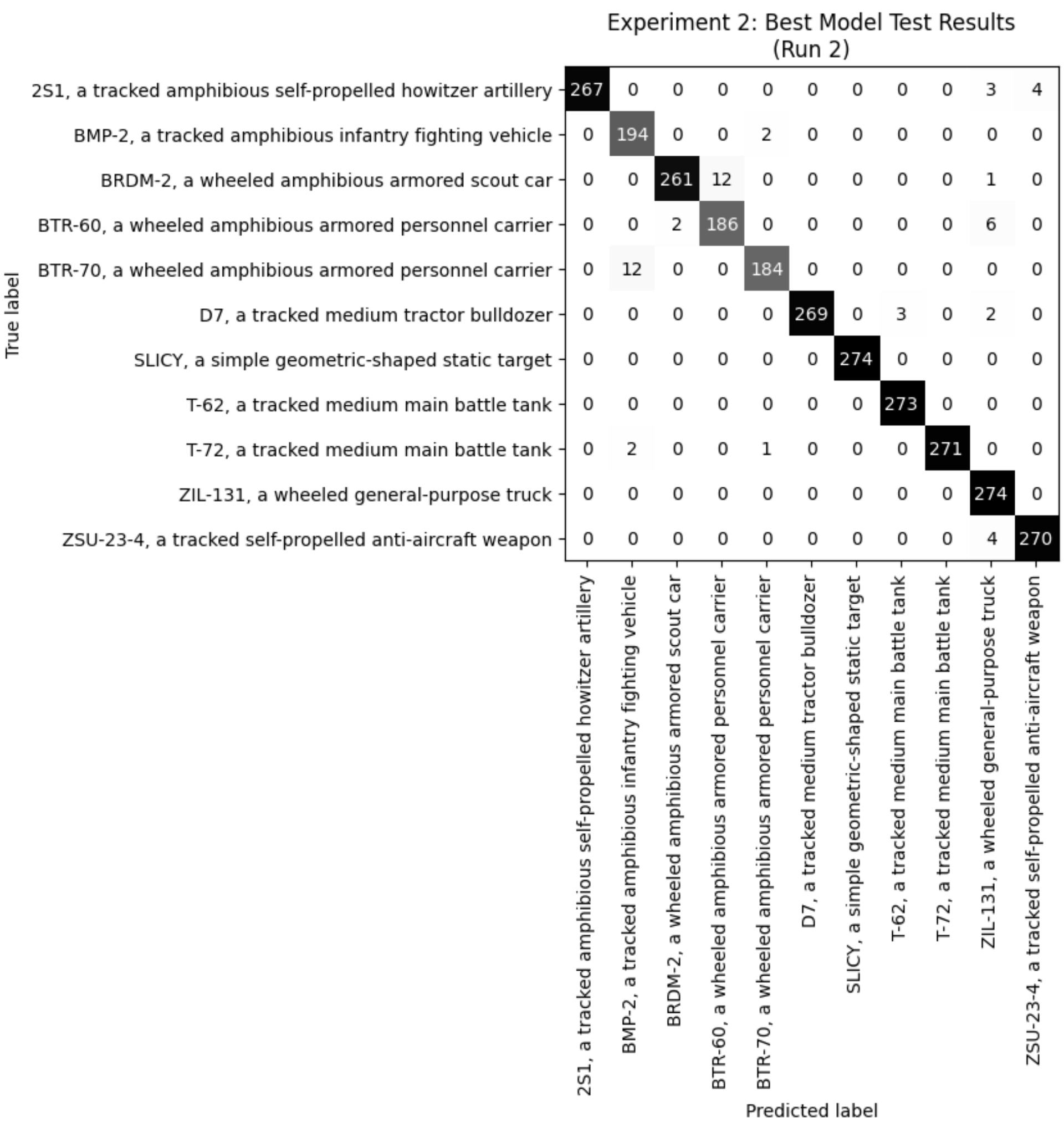

Figure 9. The simple image captioning model in Experiment 2 precisely predicts the vehicle description with no variations.

## 5. FUTURE WORK

Our results demonstrate the powerful predictive capability of an LLVM, but many open questions and opportunities remain. Although these models achieved 98% accuracy on the classification and simple image captioning test sets, further improvements are needed to fully leverage the potential of LLVM for other tasks, like VQA.

Hyperparameter tuning is a commonly used method for improving model performance. We did not conduct any hyperparameter tuning during these experiments. All parameters were set statically to default or recommended values, and we adjusted the batch size to be as large as possible for our GPU's VRAM. An interesting trend in training LLMs is the use of batch sizes comprising thousands of examples at a time to calculate a gradient, but this requires a proportionally large amount of VRAM in the GPU. We aim to identify additional methods to further enhance the training batch size. Testing different hyperparameters is typically crucial for machine learning development. Hyperparameters for training include batch size, learning rate, total epochs, warmup, as well as more substantial configurations for the optimizer, quantization, LoRA, prompt tuning, and data augmentation. We also do not use any data augmentation methods. Supplementary and synthetically generated data have been shown to improve model performance. Even when training for only a single epoch, the LLVM consistently achieves a test accuracy of over 90%. The model consistently achieved performance results approaching state-of-the-art, having only seen each training example twice.

Our LLVM's tokenization vocabulary could be improved for future experiments. As discussed in the Machine Learning Training section, the tokenization method can be adapted to use special, unallocated token IDs. This would allow vehicle names, such as the ZSU-23-4, to be represented as a single token, rather than seven tokens. This issue is shown in Figure 10. For ATR classification on these vehicle types, our LLVM must successfully predict each token in sequence from among 36,000 possible tokens. Each sequential prediction increases the possibility of error. We observe very few of these errors in our experiments; however, predicting a single token may further decrease training time, prevent overfitting, and unlock latent zero-shot predictive capabilities.

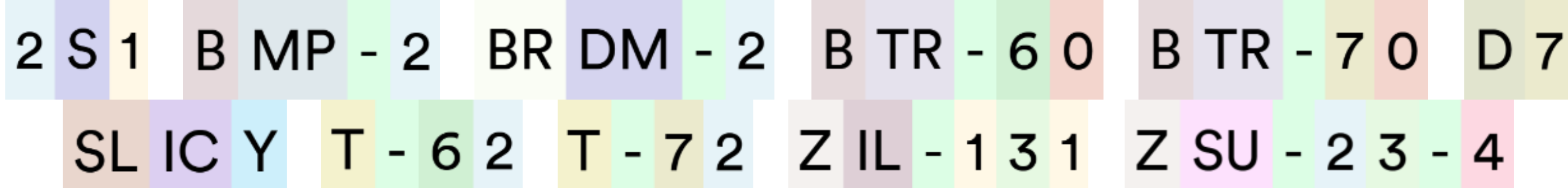

Figure 10. The target vehicle words are split by our LLVM Mistral tokenizer [27], which complicates prediction.

Recently popularized LLMs, such as the Deep Seek models, demonstrate a Chain-of-Thought reasoning capability. Such models conduct a hidden "brainstorming" before answering a posed question. This token generation process takes significantly longer to compute, but it allows the generated reasoning to be used as context for the final answer. We can leverage this Chain-of-Thought reasoning for our short-form image captioning responses. If we allow the model to generate descriptive words before predicting a vehicle classification, the LLVM can use all generated qualities towards the final assessment. In the current image captions presented in Table 2, the LLVM first makes a single challenging guess at the vehicle class, but all the following word predictions are trivial. Such sequitur predictions are not only evident to the LLVM, but they are harmful learned behavior and a symptom of overfitting.

VQA is another new research area for the MSTAR dataset. Instead of inferring the most notable information in an image, VQA learns less obvious patterns that are present in images. VQA can be used to answer challenging contextual questions, such as inferring information about subtle visual features. A VQA dataset should encompass a diverse range of question types, answers, and phrasing. Such a dataset will include many question-and-answer pairs for each SAR image. Image captioning datasets typically include at least five captions per image. We are developing a VQA benchmark challenge dataset to accompany all the MSTAR examples.

We acknowledge that there are some inconsistencies with the MSTAR data generated using the "mstar2jpeg" tool. Perhaps all the SAR images should be in the lossless TIFF data format. Unfortunately, the use of these helper functions is not widespread, with many researchers using the included JPEG image samples despite the warning by the MSTAR dataset producers. We are developing a reproducible method to create a comprehensive and consistent MSTAR image baseline.

A degree of randomness is also needed in the phrasing of both image captions, questions, and answers. When predicting the qualities of a target, randomizing the order of words and using synonyms can act as a regularization method, thereby reducing overfitting and increasing the variety of responses. Similarly, probabilistic sampling can be used to generate more flavorful and descriptive image captions. Often, in life, there is more than one correct answer. Similarly, scoring the full

SoftMax probability vector allows for more precision in scoring, as opposed to making "hard decisions" for a prediction. "Soft labels" can also be used, although this can be much more challenging to implement. During training, this method allows several optional next-tokens to be preferred, rather than just a single option.

Zero-shot prediction, open-ended question answering, and reasoning capabilities may be possible for SAR ATR with the use of an LLVM. Our future work will delve into these challenging topics in greater detail.

## 6. CONCLUSIONS

Large language and vision models are an area of rapid advancement that can be leveraged for geospatial, remote sensing, and SAR applications. Recent advances in neural network transformers, generative pretraining, LLM, and CLIP [13] are summarized in our literature review. The computational requirements of LLVM have limited access to this technology. We present our initial work to develop an LLVM for SAR ATR and the MSTAR dataset. For this narrow application, our method may be the first of its kind at the time of writing. We utilized the LLaVA-NeXT LLVM [23] architecture adapted to fit within the limits of our computer system. Quantization and LoRA are employed as PEFT methods, and all training configurations are discussed in detail for reproducibility. As part of the language model setup, ATR classification and image captioning labels are tokenized to train the LLVM, as demonstrated through visual examples. We tested the model training through two experiments, the first was vehicle classification, followed by image captions with longer descriptions. Without substantial tuning, our LLVM delivers over 98% accuracy on both tasks with the test data. We examine several model checkpoints throughout the training process, consistently demonstrating high performance. In a limited Monte Carlo experiment, several models are trained and tested with similar high performance. We conclude that LLVM models can be powerful SAR ATR predictors.

LLVM methods have progressively become more accessible for applied research. In Appendix Table 6, we examined the size and computational requirements of common LLMs, LLVMs, and their associated training strategies. We gathered the best available information describing these requirements from published literature and community notes. This table includes the quantity of GPUs, total VRAM memory, and the time required to train popular methods and foundational models of our LLVM. The specific requirements for the elements combined to form our LLVM method are noted below the table's divider. The initial ML training of the Mistral-7B LLM required approximately 500 unknown GPUs over an unreleased training time [12]. Google trained the initial ViT-Large model on 8 Tensor Processing Unit (TPU) cores [15]. This ViT-L model was then modified and further trained by OpenAI researchers for 12 days on 256 connected 32GB GPUs to create the CLIP ViT-L/14-336px vision encoder model [13]. Next, the LLaVA-NeXT researchers required 8 GPUs for 20 hours to train the combined Mistral-7B with vision encoder LLVM [23]. They first train the small 21M parameter projector, before training the entire model. Finally, we train the 4.2M LoRA parameters for our experiments. Note that we do not need to train this method from scratch or in its entirety.

Within the Results and Future Work sections, we presented shortcomings in our methodology that currently prevent zero-shot prediction on VQA tasks. Limited LoRA trainable parameters can be easily over-tuned, leading to overfitting on the MSTAR SAR images and duplicate text labels. The relatively small batch size used may produce suboptimal results. Tokenization can support more advanced uses, hyperparameter tuning is an area for exploration, and data augmentation can be highly beneficial. We must add variability to the MSTAR vehicle target captions. Experimentation is required to fully leverage the capabilities of LLVM for SAR ATR. Therefore, we are planning an MSTAR VQA benchmark dataset.

The proposed MSTAR SAR ATR LLVM yields remarkable research results and can be utilized in new ways. A SAR AI assistant could be a helpful tool for the military and intelligence communities. A trained analyst, with the assistance of agentic collaborative models, could together review SAR detections of interest. Such an AI agent could utilize advanced tools, search functions, database queries, and API requests, enabled through Chain-of-Thought reasoning, keywords, and special tokens. Teams of LLVM agents trained for specific scenarios can be organized into a management hierarchy, with workers, reviewers, coders, and report writers serving alongside professionals. The AI era is coming soon for SAR ATR.

## ACKNOWLEDGEMENTS

This work is supported in part by Prime Solutions Group Inc. and the Sensor, Signal, and Information Processing (SenSIP) Center.

## APPENDIX

Table 6. The size and training requirements of various models are presented, with the most relevant listed at the bottom.

**Model Training Requirements**

| Model | Parameters (Trainable) | VRAM Minimum | GPU Count | VRAM / GPU | VRAM Total | Training Time | GPU Hours |
|---|---|---|---|---|---|---|---|
| Transformer | 65M | 2GB | 8 x P100 | 16GB | 128GB | 12 hours | 96 |
| Transformer-L | 213M | 6.8GB | 8 x P100 | 16GB | 128GB | 84 hours | 672 |
| GPT-1 | 117M | 3.5GB | 8 x P600 | 2GB | 16GB | 720 hours | 5,760 |
| GPT-2 | 1.5B | 46.5GB | 256 x TPUv3 | 16GB | 4TB | 168 hours | 43K |
| GPT-3 | 175B | 3.1TB | 1K x V100 | 32GB | 32TB | 2K hours | 3.1M |
| GPT-4 | 1.76T | 32.4TB | 8K x A100 | 80GB | 640TB | 4K hours | 34M |
| LLaMA-1 | 65B | 1.5TB | 2048 x A100 | 80GB | 163TB | 499 hours | 1.0M |
| LLaMA-2 | 70B | 1.7TB | 2K x A100 | 80GB | 160TB | 850 hours | 1.7M |
| LLaMA-3.1 | 405B | 9.8TB | 16K x H100 | 80GB | 1,280TB | 1.9K hours | 30.84M |
| Mistral-7B | 7.3B | 178.7GB | 500 x | 80GB | 40TB | Unknown | Unknown |
| ViT-L | 307M | 9.8GB | 8 x TPUv3 | 16GB | 128GB | 720 hours | 16K |
| CLIP ViT-L/14-336 | 427.9M | 12.8GB+ | 256 x V100 | 32GB | 8.2TB | 228 hours | 1.87M |
| LLaVA-NeXT Mistral-7B | 7.6B (21M*) | 30.4GB | 8 x A100 | 80GB | 640GB | 20 hours | 160 |
| LLaVA-NeXT Mistral-7B-QLoRA (ours) | 7.6B (4.2M*) | 5.68GB | RTX3090 | 24GB | 24GB | 2.3 hours | 2.3 |

Most values are quoted from published academic research. Some values are inferred from the best available information [25].
+ CLIP also requires an LLM encoder to be aligned, significantly increasing the estimated minimum VRAM requirement.
* Only the unfrozen parameter weights require training.